\documentclass{svmult}
\pdfoutput=1

\usepackage{graphicx} 

\usepackage[bottom]{footmisc}

\usepackage{amssymb,amsmath,color,psfrag}
\usepackage{graphicx} \usepackage{xspace}
\usepackage{color}
\renewcommand{\natural}{{\mathbb{N}}} 
 \newcommand{\real}{{\mathbb{R}}}

\newcommand{\argmin}{\ensuremath{\operatorname{argmin}}}
\newcommand{\Area}{\ensuremath{\operatorname{Area}}}
 \newcommand{\map}[3]{#1:
  #2 \rightarrow #3}

\newcommand{\setdef}[2]{\{#1 \; | \; #2\}}
\newcommand{\until}[1]{\{1,\dots,#1\}}
\newcommand{\domain}{\mathcal{Q}}

\newcommand{\Topt}{T_{\mathrm{opt}}}
\newcommand{\ToptDI}{T_{\mathrm{opt}}^{\mathrm{DI}}}
\newcommand{\ToptDD}{T_{\mathrm{opt}}^{\mathrm{DD}}}

\newcommand{\SE}{\ensuremath{\operatorname{SE}}}

\newcommand{\area}{\mathcal{A}}
\newcommand{\mintime}{\tau}

\newcommand{\vmax}{v_{\text{max}}}
\newcommand{\umax}{u_{\text{max}}}
\newcommand{\wmax}{w_{\text{max}}}
\newcommand{\weberfunc}[2]{\mathcal{H}_{#1}^*(#2)}

\newcommand{\reach}{\mathcal{R}}
\newcommand{\TMSDI}{T_{\mathrm{MS}}^{\mathrm{DI}}}
\newcommand{\TMSDD}{T_{\mathrm{MS}}^{\mathrm{DD}}}
\newcommand{\TSLDI}{T_{\mathrm{SL}}^{\mathrm{DI}}}
\newcommand{\TMCDD}{T_{\mathrm{MC}}^{\mathrm{DD}}}
\newcommand{\rhostar}{\rho^*}
\newcommand{\vstar}{v^*}
\newcommand{\todo}[1]{\par\noindent{\color{red}\raggedright\sc{#1}\par\marginpar{\Large \bf $\star$}}}
\renewcommand{\todo}[1]{}

\begin{document}

\title*{On Endogenous Reconfiguration in Mobile Robotic Networks}

%
\author{Ketan Savla\inst{1} \quad Emilio Frazzoli\inst{1}}

\institute{Laboratory for Information and Decision Systems,
  Massachusetts Institute of Technology \texttt{\{ksavla,frazzoli\}@mit.edu} }

\maketitle

\setcounter{footnote}{0}

\renewcommand{\abstractname}{Abstract:} 
\begin{abstract}
In this paper, our focus is on certain applications for mobile robotic networks, where reconfiguration is driven by factors intrinsic to the network rather than changes in the external environment. In particular, we study a version of the coverage problem useful for surveillance applications, where the objective is to position the robots in order to minimize the average distance from a random point in a given environment to the closest robot. This problem has been well-studied for omni-directional robots and it is shown that optimal configuration for the network is a centroidal Voronoi configuration and that the coverage cost belongs to $\Theta(m^{-1/2})$, where $m$ is the number of robots in the network. In this paper, we study this problem for more realistic models of robots, namely the double integrator (DI) model and the differential drive (DD) model. We observe that the introduction of these motion constraints in the algorithm design problem gives rise to an interesting behavior. For a \emph{sparser} network, the optimal algorithm for these models of robots mimics that for omni-directional robots. We propose novel algorithms whose performances are within a constant factor of the optimal asymptotically (i.e., as $m \to +\infty$). In particular, we prove that the coverage cost for the DI and DD models of robots is of order $m^{-1/3}$. Additionally, we show that, as the network grows, these novel algorithms outperform the conventional algorithm; hence necessitating a reconfiguration in the network in order to maintain optimal quality of service. 
\end{abstract}

\section{Introduction}
The advent of large scale sensor and robotic networks has led to a surge of interest in reconfigurable networks. These systems are usually designed to reconfigure in a \emph{reactive} way, i.e., as a response to changes in external conditions. Due to their importance in sensor network applications, reconfiguration algorithms have attracted a lot of attention, e.g., see \cite{Kansal.Kaiser.ea:TOSN07}. However, there are very few instances in engineering systems, if any, that demonstrate an internal reconfiguration in order to maintain a certain level of performance when certain \emph{intrinsic} properties of the system are changed. However, examples of endogenous reconfiguration or phase transitions are abound in nature, e.g., desert locusts~\cite{Buhl.Sumpter.ea:06} who switch between gregarious and social behavior abruptly, etc. An understanding of the phase transitions can not only  provide insight into the reasons for transitions in naturally occurring systems but also identify some design principles involving phase transition to maintain efficiency in engineered systems.

In this paper, we observe such a phenomenon under a well-studied setting that is relevant for various surveillance applications. We consider a version of the  so-called Dynamic Traveling Repairperson Problem, first  proposed by \cite{Psaraftis:88} and later developed in \cite{Bertsimas.vanRyzin:91}. In this problem, service requests are generated dynamically. In order to fulfill a request, one of the vehicles needs to travel to its location. The objective is to design strategies for task assignment and motion planning of the robots that minimizes the average waiting time of a service request. In this paper, we consider a special case of this problem when service requests are generated sparingly. This problem, also known as coverage problem, has been well-studied in the robotics and operations research community. However, we consider the problem in the context of realistic models of robots: double integrator models and differential drive robots. Some preliminary work on coverage for curvature-constrained vehicles was reported in our earlier work~\cite{Enright.Savla.ea:CDC08}. In this paper, we observe that when one takes into consideration the motion constraints of the robots, the optimal solution exhibits a phase transition that depends on the size of the network. 

The contributions of this paper are threefold. First, we identify an interesting characteristic of the solution to the coverage problem for double integrator and differential drive robots, where reconfiguration is necessitated intrinsically by the growth of the network, in order to maintain optimality. Second, we propose novel approximation algorithms for double integrator robots as well as differential drive robots and prove that they are within a constant factor of the optimal in the asymptotic ($m \to +\infty$) case. Moreover, we prove that, asymptotically, these novel algorithms will outperform the conventional algorithms for omni-directional robots. Lastly, we show that the coverage for both, double integrator as well as differential drive robots scales as $1/m^{1/3}$ asymptotically.

\section{Problem Formulation and Preliminary Concepts}
In this section, we formulate the problem and present preliminary concepts.

\subsection*{Problem Formulation}
The problem that we consider falls into the general category of the so called \emph{Dynamic Traveling Repairperson Problem}, originally proposed in \cite{Psaraftis:88}.
 Let $\domain \subset \real^2$
be a convex, compact domain on the plane, with non-empty interior; we
will refer to $\domain$ as the {\em environment}.  For simplicity in presentation, we assume that $\domain$ is a square, although all the analysis presented in this paper carries through for any convex and compact $\domain$ with non-empty interior in $\real^2$.  Let $\area$ be
the area of $\domain$.  A spatio-temporal Poisson process generates
{\em service requests} with finite time intensity $\lambda > 0$ and uniform
spatial density inside the environment. In this paper, we focus our attention on the case when $\lambda \to 0^+$, i.e., when the service requests are generated very rarely.
These service
requests are identical and are to be fulfilled by a team of $m$
robots. A service request is fulfilled when one of $m$ robots moves to the
target point associated with it. 
 
 We will consider two robot models: the double integrator model and the differential drive model.
The double integrator (DI) model describes the dynamics of 
a robot with significant inertia. The configuration 
of the robot is $g = (x, y, v_x,v_y) \subset 
\real^4$ where $(x, y)$ is the position of the robot in Cartesian 
coordinates, and $(v_x,v_y)$ is its velocity. The dynamics of the 
DI robot are given by
\begin{align*}
\dot{x}(t) &=v_x(t),\\
\dot{y}(t) &= v_y(t), \quad v_x(t)^2+v_y(t)^2 \leq \vmax^2 \quad \forall t\\
\dot{v}_x(t) & = u_x(t),\\
\dot{v}_y(t) &= u_y(t), \quad u_x(t)^2+u_y(t)^2 \leq \umax^2 \quad \forall t,
\end{align*}
where $\vmax$ and $\umax$ are the bounds on the speed and the acceleration of the robots. 

The differential drive model describes the kinematics of 
a robot with two independently actuated wheels, each a 
distance $\rho$ from the center of the robot. The configuration 
of the robot is a directed point in the plane, $g = (x, y, \theta) \subset 
\SE(2)$ where $(x, y)$ is the position of the robot in Cartesian 
coordinates, and $\theta$ is the heading angle with respect to the $x$ 
axis. 
 The dynamics of the DD robot are given by
\begin{align*}
\dot{x}(t)&=\frac{1}{2}(w_l(t)+w_r(t)) \cos \theta(t),\\
\dot{y}(t)&=\frac{1}{2}(w_l(t)+w_r(t)) \sin \theta(t),\\
\dot{\theta}(t)&=\frac{1}{2 \rho} (w_r(t)-w_l(t)), \quad |w_l(t)| \leq \wmax \forall t, |w_r(t)| \leq \wmax \forall t, 
\end{align*}
where the inputs $w_l$ and $w_r$ are the angular velocities of the left and the right wheels, which we assume to be bounded by $\wmax$. Here, we have also assumed that the robot wheels have unit radius.

The robots are assumed to be identical. The strategies of the robots in the presence and absence of service requests are governed by their \emph{motion coordination algorithm}.
A motion coordination algorithm is a function that determines the actions
of each robot over time. For the time being, we will denote these
functions as $\pi = (\pi_1, \pi_2, \ldots, \pi_m)$, but do not
explicitly state their domain; the output of these functions is a
steering command for each vehicle. The objective is the design of
motion coordination algorithms that allow the robots to fulfill
service requests efficiently. To formalize the notion of efficiency, let $T_j$ be the time
elapsed between the generation of the $j$-th service request, and the
time it is fulfilled and let $T_\pi:=  \lim_{j\to +\infty}  \lim_{\lambda \to 0^+} \mathrm{E}[T_j]$ 
\todo{do you need to state the $\lambda$ limit here? $T_\pi$ can be defined for all values of $\lambda$, right? (Assuming stability)}
be defined as the system time under policy $\pi$, i.e., the expected time a
service request must wait before being fulfilled, given that the
robots follow the algorithm defined by $\pi$. We shall also refer to the average system time as the \emph{coverage cost}. Note that the system
time $T_\pi$ can be thought of as a measure of the quality
of service collectively provided by the robots.

In this paper, we wish to devise motion coordination
algorithms that yield a quality of service
(i.e., system time) achieving, or approximating, the theoretical
optimal performance given by $\Topt = \inf_{\pi} T_\pi$. Since finding the optimal algorithm maybe computationally intractable, we are also interested in designing computationally efficient algorithms
that are within a constant factor of the optimal, i.e.,
policies $\pi$ such that $T_\pi \le \kappa
\Topt$ for some constant $\kappa$. Moreover, we are interested in studying the scaling of the performance of the algorithms with $m$, i.e., size of the network, other parameters remaining constant. Since, we keep $\area$ fixed, this is also equivalent to study the scaling of the performance with respect to the density $m/\area$ of the network.

We now describe how a solution to this problem gives rise to an endogenous reconfiguration in the robotic network.

\subsection*{Endogenous Reconfiguration}
The focus of this paper is on \emph{endogenous} reconfiguration, that is a reconfiguration necessitated by the growth of the network (as the term `endogenous' implies in biology), as opposed to any external stimulus. We formally describe its meaning in the context of this paper.
In the course of the paper, we shall propose and analyze various algorithms for the coverage problem. In particular, for each model of the robot, we will propose two algorithms, $\pi_1$ and $\pi_2$. The policy $\pi_1$ closely resembles the omni-directional based policy, whereas the novel $\pi_2$ policy optimizes the performance when the motion constraints of the robots start playing a significant role. We shall show that 
\begin{equation*}
\lim_{m/\area \to 0^+} \frac{T_{\pi_1}}{\Topt}  =1,\quad \lim_{m \to +\infty} \frac{T_{\pi_2}}{T_{\pi_1}}  = 0,\quad \limsup_{m \to + \infty}  \frac{T_{\pi_2}}{\Topt}  \leq c \text{ for some constant $c>1$}.
\end{equation*}
\todo{Limit $m \to 0^+$? Wouldn't it be $m \to 1$, with $m$ integer greater than 1? If you say that $m$ is a positive integer, then you can get away with the `+'s, i.e., you can just write $\lim_{m \to 1}$ and $\lim_{m \to \infty}$ without confusions}
This shows that, for sparse networks, the omni-directional model based algorithm $\pi_1$ is indeed a reasonable algorithm. However, as the network size increases, there is a phase transition, during which the motion constraints start becoming important and the $\pi_2$ algorithm starts outperforming the $\pi_1$ algorithm. Moreover, the $\pi_2$ algorithm performs with a constant factor of the optimal in the asymptotic ($m \to +\infty$) case. Hence, in order to maintain efficiency, one needs to switch away from the $\pi_1$ policy as the network grows. 
It is in this sense that we shall use the term endogenous reconfiguration to denote a switch in the optimal policy with the growth of the network. 

\section{Lower Bounds}
In this section, we derive lower bounds on the coverage cost for the robotic network that are independent of any motion coordination algorithm adopted by the robots. Our first lower bound is obtained by modeling the robots as \emph{equivalent} omni-directional robots. Before stating the lower bounds formally, we need to briefly review a related problem from computational geometry which has direct consequences for the case omni-directional robots.

\subsection*{The Continuous $m$-median Problem}
Given a convex, compact set
$\domain\subset\real^2$ and a set of points $p = \{p_1, p_2, \ldots,
p_m\} \in \domain^m$, the expected distance between a random point
$q$, sampled from a uniform distribution over $\domain$, and the
closest point in $p$ is given by
\begin{equation*}
\label{eq:median}
\mathcal{H}_m(p,\domain) :=\; \int_\domain\frac{1}{\area} \min_{i \in \{1,\ldots,m\}} \|p_i-q\| \ dq= 
\sum_{i=1}^m\int_{\mathcal{V}_i(p)} \frac{1}{\area} \|p_i-q\| \ dq,
\end{equation*}
where $\mathcal{V}(p) = (\mathcal{V}_1(p), \mathcal{V}_2(p), \ldots,
\mathcal{V}_m(p))$ is the {\em Voronoi (Dirichlet)
  partition}~\cite{Okabe.Boots.ea:00} of $\domain$ generated by the
points in $p$, i.e., $$\mathcal{V}_i(p) = \{q \in \domain: \|q-p_i\|
\le \|q-p_j\|, \forall j \in \{1, \ldots, m\} \},\qquad i \in \{1,
\ldots, m\}.$$ The problem of choosing $p$ to minimize $\mathcal{H}_m$ is known
in geometric optimization~\cite{Agarwal.Sharir:98} and facility
location~\cite{Drezner:95} literature as the (continuous) $m$-median problem. The
$m$-median of the set $\domain$ is the global minimizer
\begin{equation*}
  p^*_m(\domain) = \argmin_{p \in \domain^m} \mathcal{H}_m(p,\domain).
\end{equation*}
We let $\weberfunc{m}{\domain} = \mathcal{H}_m(p^*_m(\domain),\domain)$ be the global
minimum of $\mathcal{H}_m$. 
The solution of the continuous $m$-median problem is hard in the general
case because the function $p\mapsto
\mathcal{H}_m(p,\domain)$ is not convex for $m>1$. 
 However, gradient algorithms for the continuous multi-median
problem can be designed~\cite{Cortes.Martinez.ea:04}. We would not go further into the details of computing these $m$-median locations and assume that these locations are given or that a computationally efficient algorithm for obtaining them is available.

This particular problem formulation, with demand generated
independently and uniformly from a continuous set, is studied
thoroughly in \cite{Papadimitriou:81} for square regions and
\cite{Zemel:84} for more general compact regions.  It is shown in
\cite{Zemel:84} that, in the asymptotic ($m \rightarrow +\infty$)
case, $\weberfunc{m}{\domain} = c_\mathrm{hex} \sqrt{\frac{\area}{m}} \ \ \
\mbox{almost surely,}$ where $c_\mathrm{hex} \approx 0.377$ is the
first moment of a hexagon of unit area about its center.  This optimal
asymptotic value is achieved by placing the $m$ points on a regular
hexagonal network within $\domain$ (the honeycomb heuristic).  Working
towards the above result, it is also shown in \cite{Zemel:84} that for
any $m \in \natural$:
\begin{equation}
\label{eq:weber-bounds}
\frac{2}{3} \sqrt{\frac{\area}{\pi m}} \le \weberfunc{m}{\domain} \le  c(\domain)\sqrt{\frac{\area}{m}},
\end{equation}
where $c(\domain)$ is a constant depending on the shape of $\domain$. In particular, for a square $\domain$, $c(\domain) \approx 0.38$. 



We use two different superscripts on $\Topt$ for the two models of the robots, i.e., we use $\ToptDI$ for the DI robot and $\ToptDD$ for the DD robot. Finally, we state a lower bound on these quantities as follows.

\begin{lemma}
\label{lem:lower-bound-trivial}
The coverage cost satisfies the following lower bound.
\begin{equation*}
\ToptDI \geq \frac{\weberfunc{m}{\domain}}{\vmax}, \quad \ToptDD \geq \frac{\weberfunc{m}{\domain}}{\wmax}.
\end{equation*}
\end{lemma}
\begin{proof}
The proof follows trivially by relaxing the constraints on the robots and allowing them to move like omni-directional robots with speeds $\vmax$ for DI robots and $\wmax$ for DD robots. One can then adopt the lower bound on coverage cost for omni-directional robot, e.g., \cite{Bertsimas.vanRyzin:91} to arrive at the result.
\end{proof}

\begin{remark}
Since from Equation~\eqref{eq:weber-bounds}, $\weberfunc{m}{\domain} \in \Omega(1/\sqrt{m})$, Lemma~\ref{lem:lower-bound-trivial} implies that $\ToptDI$ and $\ToptDD$ also belong to $\Omega(1/\sqrt{m})$.
\end{remark}

This lower bound will be particularly useful for proving the optimality of algorithms for sparse networks. We now proceed towards deriving a tighter lower bound which will be relevant for dense networks. The reachable sets of the two models of robots will play a crucial role in deriving the new lower bound. We study them next.

\subsection*{Reachable Sets for the Robots}
In this subsection, we state important properties of the reachable sets of the double integrator and differential drive robots that are useful in obtaining tighter lower bound.

Let $\map{\mintime}{G \times \real^2}{\real^+}$ be the minimum time required to steer a robot from initial configuration $g$ in $G$ to
a point $q$ in the plane. For the DI robot, $G=\real^4$ and for DD robots $G=\SE(2)$. With a slight abuse of terminology, we define the reachable set of a robot, $\reach_t(g)$, to be the set of all points $q$ in $\domain$ that are reachable in time $t>0$ starting at configuration $g$. Note that, in this definition, we do not put any other constraint (e.g., heading angle, etc.) on the terminal point $q$. Formally, the reachable set is defined as
\begin{equation*}
\reach_t(g)=\setdef{q \in \real^2}{\mintime(g,q) \leq t}.
\end{equation*}

We now state a series of useful properties of the reachable sets.
\begin{lemma}[Upper bound on the small-time reachable set area]
\label{lem:reach-set-area-upper-bound}
The area of the reachable set for a DI robot starting at a configuration $g=(x,y,\dot{x},\dot{y}) \in \real^4$, with $\dot{x}^2+\dot{y^2} = v_0$, satisfies the following upper bound. 
\begin{equation*}
\Area(\reach_t(g)) \leq \left\{\begin{array}{ll} 
  2 v_0 \umax t^3 + o(t^3), \quad \text{as $t \to 0^+$}
    & \mbox{for } v_0 > 0,\\
   \umax^2 t^4 & \mbox{for } v_0=0.
\end{array}  \right.
\end{equation*}
The area of the reachable set for a DD robot starting at any configuration $g \in \SE(2)$ satisfies the following upper bound. 
\begin{equation*}
\Area(\reach_t(g)) \leq \frac{5}{6 \rho} \wmax^3 t^3 + o(t^3), \quad \text{as $t \to 0^+$}.
\end{equation*}
\end{lemma}

\begin{proof}
The result for the differential drive robot has been derived in \cite{Enright.Frazzoli:CDC06}.
We derive the result for the double integrator robot here. Assume, without any loss of generality, that the robot is initially placed at the origin with velocity aligned with the $x$-axis. The maximum of the absolute value of the $x$-coordinate and $y$-coordinate of all the points reachable in time $t$ is less than or equal to $v_0 t+\frac{1}{2}\umax t^2$ and $\frac{1}{2}\umax t^2$, respectively. Therefore, the area of the reachable set is trivially upper bounded by $4(v_0 t+\frac{1}{2}\umax t^2)(\frac{1}{2}\umax t^2)=2v_0 \umax t^3 + \umax^2 t^4$.
\end{proof}

\begin{lemma}[Lower bound on the reachable set area]
\label{lem:reach-set-area-lower-bound}
The area of the reachable set of a DI robot starting at a configuration $g=(x,y,\dot{x},\dot{y}) \in \real^4$, with $\dot{x}^2+\dot{y}^2=v_0$, satisfies the following lower bound. 
\begin{equation*}
\Area(\reach_t(g)) \geq \frac{v_0 \umax}{3} t^3 \quad \forall t \leq \frac{\pi}{2} \frac{v_0^2}{\umax}.
\end{equation*}
The area of the reachable set of a DD robot starting at any configuration $g \in \SE(2)$ satisfies the following lower bound. 
\begin{equation*}
\Area(\reach_t(g)) \geq \frac{2}{3 \rho} \wmax^3 t^3.
\end{equation*}
\end{lemma}

\begin{lemma}
\label{lem:travel-time-integral}
The travel time to a point in the reachable set for a DI robot starting at a configuration $g=(x,y,\dot{x},\dot{y}) \in \real^4$, with $\dot{x}^2+\dot{y}^2=v_0$, satisfies the following property.
\begin{equation*}
\int_{\reach_t(g)} \tau(g,q) dq \geq \frac{v_0 \umax}{12} t^4 \quad \forall t \leq \frac{\pi}{2} \frac{v_0^2}{\umax}.
\end{equation*} 
The travel time to a point in the reachable set for a DD robot starting at any configuration $g \in \SE(2)$ satisfies the following property.
\begin{equation*}
\int_{\reach_t(g)} \tau(g,q) dq \geq \frac{\wmax^3}{6 \rho} t^4.
\end{equation*} 
\end{lemma}	

\begin{proof}
For both the robots, $\int_{\reach_t(g)} \tau(g,q) dq=\int_0^t \Area(\reach_s(g)) ds$. Using Lemma~\ref{lem:reach-set-area-lower-bound}, for a double integrator robot, for all $t \leq \frac{\pi}{2} \frac{v_0^2}{\umax}$ we have that,
\begin{equation*}
\int_{\reach_t(g)} \tau(g,q) dq \geq \frac{v_0 \umax}{3} \int_0^t s^3 ds=  \frac{v_0 \umax}{12} t^4.
\end{equation*}
The proof for the differential drive robot follows along similar lines.
\end{proof}

We are now ready to state a new lower bound on the coverage cost.

\begin{theorem}[Asymptotic lower bound on the coverage cost]
\label{thm:asymp-lower-bound}
The coverage cost for a network of DI or DD robots satisfies the following asymptotic lower bound.
 \begin{align*}
 \liminf_{m \to + \infty} \ToptDI m^{1/3} & \geq \frac{1}{24} \Big( \frac{\area}{2 \vmax \umax}\Big)^{1/3}, \quad \text{and} \\  \liminf_{m \to + \infty} \ToptDD m^{1/3} & \geq \frac{1}{5 \wmax} \Big(\frac{6 \rho \area}{5}\Big)^{1/3}.
 \end{align*}
\end{theorem}
\begin{proof}
We state the proof for the double integrator robot. The proof for the differential drive robot follows along similar lines.

In the following, we use the notation $\mathrm{A}_i = \mathrm{Area}(\mathcal{DV}_i(g))$, where $\mathcal{DV}_i(g):=\setdef{q \in \domain}{\tau(g_i,q) \leq \tau(g_j,q) \quad \forall j \neq i}$.  We begin with
\begin{align}
  \ToptDI & = \inf_{g \in \real^{4m}} \sum_{i}^m \int_{\mathcal{DV}_i(g)} \frac{1}{\area} \tau(g_i,q)\ dq \nonumber \\
  & \ge \inf_{g \in \real^{4m}} \int_\domain\frac{1}{\area}\min_{i \in \{1,\ldots, m\}}  \tau(g_i,q)dq.
\label{eq:ToptDI-step0}
\end{align}
Let $\bar{R}_{\mathrm{A}_i}(g_i)$ be the reachable set starting at configuration $g$ and whose area is $\mathrm{A}_i$. Using the fact that, given an area $\mathrm{A}_i$, the region with the minimum integral of the travel time to the points in it is the reachable set of area $\mathrm{A}_i$, one can write Equation~\eqref{eq:ToptDI-step0} as
\begin{equation}
\ToptDI  \ge  \inf_{g \in \real^{4m}} \sum_{i}^m \int_{\bar{R}_{\mathrm{A}_i}(g_i)}  \frac{1}{\area} \tau(g_i,q)\ dq.
\label{eq:ToptDI-step1}
\end{equation} 
Let $t_i$ be defined such that $\Area(\reach_{t_i}(g_i))=\mathrm{A}_i$. Lets assume that as $m \to +\infty$, $\mathrm{A}_i \to 0^+$ (this point will be justified later on). In that case, we know from Lemma~\ref{lem:reach-set-area-upper-bound} that, $t_i$ can be lower bounded as $t_i \geq \Big( \frac{\mathrm{A}_i}{2 v_i \umax}\Big)^{1/3}$, where $v_i$ is the speed associated with the state $g_i$. Therefore, from Equation~\eqref{eq:ToptDI-step1} and Lemma~\ref{lem:travel-time-integral}, one can write that 
\begin{align}
 \ToptDI  \geq & \inf_{g \in \real^{4m}} \sum_{i}^m \int_{\reach_{t_i}(g_i)}  \frac{1}{\area} \tau(g_i,q)\ dq \nonumber \\
 \geq &  \inf_{g \in \real^{4m}} \sum_{i}^m  \frac{1}{\area} \frac{v_i \umax}{12} t_i^4.
 \label{eq:ToptDI-step2}
\end{align}
Using the above mentioned lower bound on $t_i$, Equation~\eqref{eq:ToptDI-step2} can be written as
\begin{align*}
 \ToptDI  \geq & \frac{1}{24 \sqrt[3]{2} \area} \frac{1}{\umax^{1/3}} \inf_{g \in \real^{4m}} \frac{\mathrm{A}_i^{4/3}}{v_i^{1/3}}\\
 \geq & \frac{1}{24 \sqrt[3]{2} \area} \frac{1}{\umax^{1/3} \vmax^{1/3}} \min_{\{\mathrm{A}_1,\mathrm{A}_2,\dots,\mathrm{A}_m \} \in \real^{m}} \sum_{i}^m \mathrm{A}_i^{4/3}\\
 &  \mathrm{subject} \ \mathrm{to}\ \ \ \sum_i^m\mathrm{A}_i =
  \area \quad \text{and } \mathrm{A}_i \geq 0 \quad \forall i \in
  \until{m}.
\end{align*}
Note that the function $f(x)=x^{4/3}$ is continuous, strictly
increasing and convex. Thus by using the Karush-Kuhn-Tucker conditions
\cite{Boyd.Vandenberghe:04}, one can show that the quantity
$\sum_{i}^m \mathrm{A}_i^{4/3}$ is minimized with an equitable partition, i.e., $\mathrm{A}_i =
\area/m,\ \forall i$. This also justifies the assumption that for $m \to +\infty$, $\mathrm{A}_i \to 0^+$.
\end{proof}

\begin{remark}
Theorem~\ref{thm:asymp-lower-bound} shows that  both $\ToptDI$ and $\ToptDD$ belong to $\Omega(1/m^{1/3})$.
\end{remark}

\section{Algorithms and their Analyses}
In this section, we propose novel algorithms, analyze their performance and explain how the size of the network plays a role in selecting the right algorithm.

We start with an algorithm which closely resembles the one for omni-directional robots. Before that, we first make a remark relevant for the analysis of all the algorithms to follow.

\begin{remark}
Note that if $n_0$ is the number of outstanding service requests at initial time, then the time required to service all of them is finite ($\domain$ being bounded). During this time period, the probability of appearance of a new service requests appearing is zero (since we are dealing with the case when the rate of generation of targets $\lambda$ is arbitrarily close to zero). Hence, after an initial transient, with probability one, all the robots will be in their stationary locations at the appearance of the new target. Moreover, the probability of number of outstanding targets being more than one is also zero. Hence, in the analysis of the algorithms, without any loss of generality, we shall implicitly assume that there no outstanding service requests initially.
\end{remark}

\subsection*{The Median Stationing (MS) Algorithm}
\begin{figure}[htb]
 \begin{minipage}[c]{.49\textwidth}
  \centerline{\includegraphics[width=0.9\linewidth]{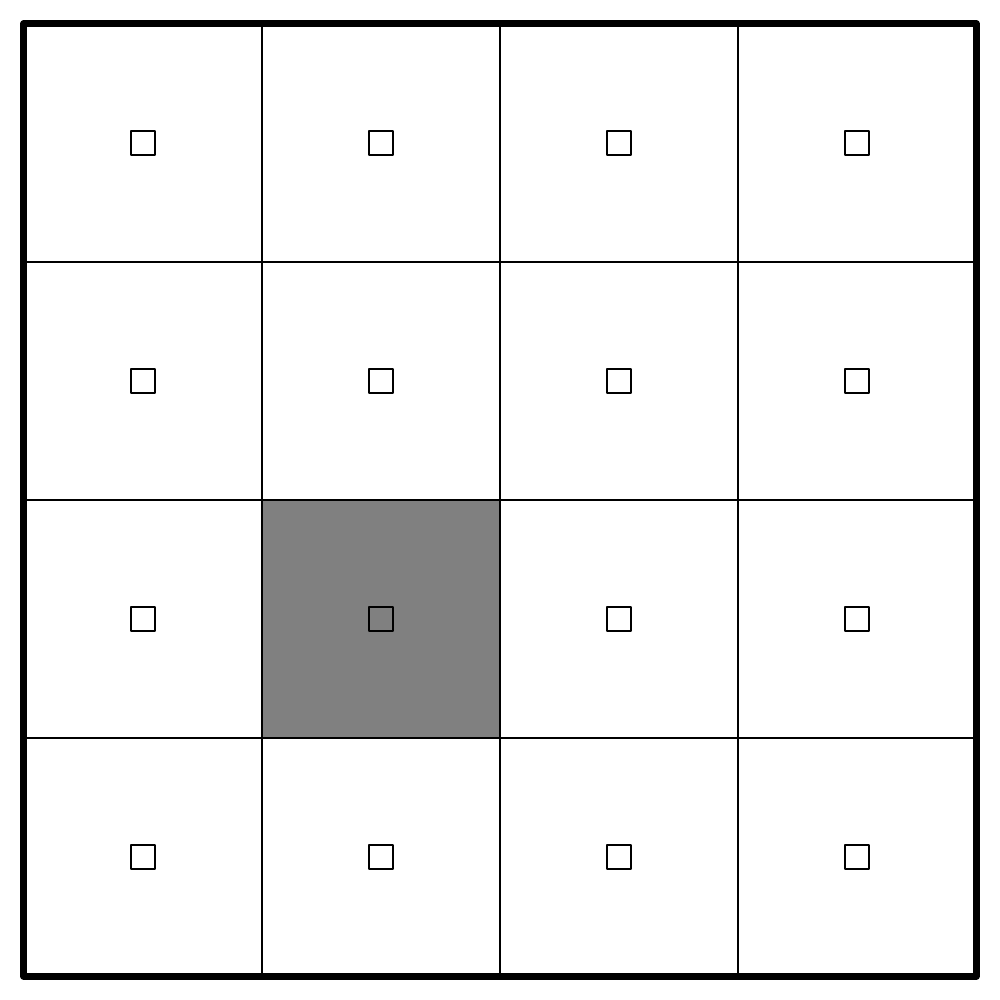}}
 \end{minipage}
 \begin{minipage}[c]{.49\textwidth}
  \centerline{\includegraphics[width=0.9\linewidth]{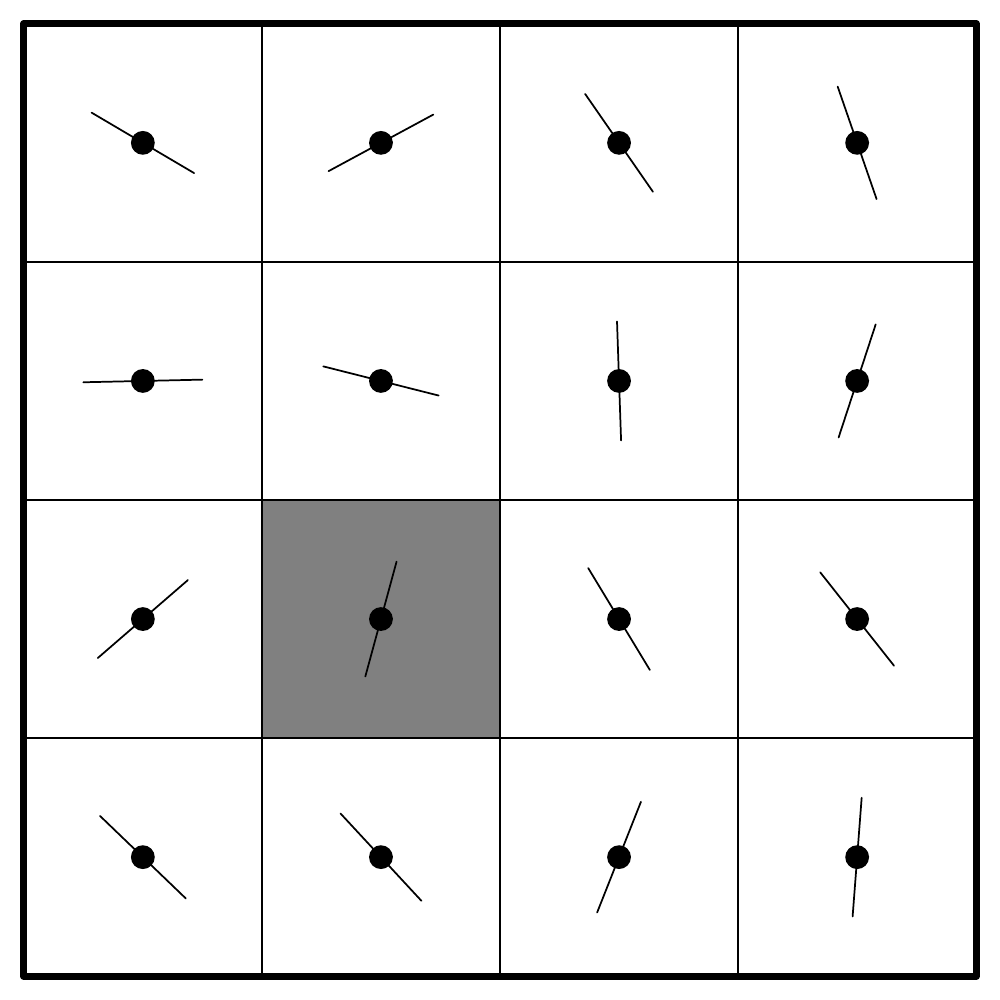}}
 \end{minipage}
 \caption{Depiction of typical stationary configurations for the Median Stationing Algorithm. On the left: DI robots at rest at their stationary locations. On the right: DD robots with arbitrary headings at their stationary locations. In both the figures, the shaded cell represents a typical \emph{region of responsibility} for a robot.}
 \label{fig:medianstation}
 \end{figure}

Place the $m$ robots at rest at the $m$-median locations of $\domain$. In case of the DD robot, its heading is chosen arbitrarily. These $m$-median locations will be referred to as the \emph{stationary locations} of the robots. When a service request appears, it is assigned to the robot whose stationary location is closest to the location of the service request. In order to travel to the service location, the robots use the fastest path with no terminal constraints at the service location. In absence of outstanding service tasks, the robot returns to its stationary location. The stationary configurations are depicted in Figure~\ref{fig:medianstation}. 

Let $\TMSDI$ and $\TMSDD$ be the coverage cost as given by the above policy for the DI and DD robots, respectively. 

\begin{theorem}[Analysis of the MS algorithm]
\label{thm:MS-bound}
The coverage cost for a network of DI and DD robots with the MS algorithm satisfies the following bounds.
\begin{equation*}
 \frac{\weberfunc{m}{\domain}}{\vmax} \leq \TMSDI \leq \frac{\weberfunc{m}{\domain}}{\vmax}+ \frac{\vmax}{2 \umax} +  \sqrt{\frac{2 \sqrt{2 \area}}{\umax}}.
\end{equation*} 
\begin{equation*}
 \frac{\weberfunc{m}{\domain}}{\wmax} \leq \TMSDD \leq \frac{\weberfunc{m}{\domain}}{\wmax}+\frac{\rho \pi}{2 \wmax}.
\end{equation*} 
\end{theorem} 
\begin{proof}
The lower bounds on $\TMSDI$ and $\TMSDD$ follow trivially from Lemma~\ref{lem:lower-bound-trivial}.

For a double integrator robot, the minimum travel time from rest at location $p$ to a point $q$ is given by 
\begin{equation*}
\tau((p,0),q) \leq \left\{\begin{array}{ll} 
  \sqrt{\frac{2 \|p-q\|}{\umax}}
    & \mbox{for } \|p-q\| \leq \frac{\vmax^2}{2 \umax},\\
  \frac{\|p-q\|}{\vmax} + \frac{\vmax}{2 \umax} & \mbox{otherwise}.
\end{array}  \right.
\end{equation*}
Therefore, for any $q$, $\tau((p,0),q)$ can be upper bounded as
\begin{align*}
\tau((p,0),q) \leq & \frac{\|p-q\|}{\vmax} + \frac{\vmax}{2 \umax} +  \sqrt{\frac{2 \|p-q\|}{\umax}} \\
\leq &  \frac{\|p-q\|}{\vmax} + \frac{\vmax}{2 \umax} +  \sqrt{\frac{2 \sqrt{2 \area}}{\umax}}.
\end{align*}
The upper bound on $\TMSDI$ is then obtained by taking the expected value over all $q \in \domain$ while taking into consideration the assignment policies for the services to robots.
For a differential drive robot, the travel time from any initial configuration $(p,\theta)$ to a point $q$ can be upper bounded by $\frac{\|p-q\|}{\wmax}+\frac{\pi \rho}{2 \wmax}$.  The result follows by taking expected value of the travel time over all points in $\domain$.
\end{proof}

\begin{remark}
Theorem~\ref{thm:MS-bound} and Lemma~\ref{lem:lower-bound-trivial} along with Equation~\eqref{eq:weber-bounds} imply that,
$$\lim_{m/\area \to 0^+} \frac{\TMSDI}{\ToptDI}=1 \quad \text{and} \quad \lim_{m/\area \to 0^+} \frac{\TMSDD}{\ToptDD}=1.$$
This implies that the MS algorithm is indeed a reasonable algorithm for sparse networks, where the travel time for a robot to reach a service location is almost the same as that for an omni-directional vehicle.

However, as the density of robots increases, the assigned service locations to the robots start getting relatively closer. In that case, the motion constraints start having a significant effect on the travel time of the robots and it is not obvious in that case that the MS algorithm is indeed the best one.
\end{remark}

In fact, for dense networks, one can get a tighter lower bound on the performance of the MS algorithm: 
\begin{theorem}
\label{thm:tighter_lb}
The coverage cost of a dense network of DI and DD robots with the MS algorithm satisfies the following bounds: 
\begin{equation*}
\liminf_{m \to \infty}  \TMSDI m^{1/4}\geq 
0.321 \left(\frac{\mathcal{A}}{\umax^2}\right)^{1/4}.
\end{equation*}
\begin{equation*}
\liminf_{m \to \infty}  \TMSDD \geq 
\frac{\pi\rho}{4\wmax}.
\end{equation*}
\end{theorem}

\begin{proof}
Let us consider the DI case first. 
The minimum travel time for the $i$-th robot to reach a point $q$ inside 
its region of responsibility $\mathcal{V}_i$, starting from rest at the median location $p_i$ is bounded by 
\begin{equation}
\label{eq:bound_acc}
\tau((p_i,0), q) \ge \max \left\{ \frac{\|p_i-q\|}{\sqrt{2 \umax d_{\mathcal{V}_i}}}, \frac{\|p_i-q\|}{\vmax} \right\},
\end{equation}
where 
$$d_{\mathcal{V}_i} = \max_{q \in \mathcal{V}_i} \|p_i-q\|.$$
For large $m$, it is known that the honeycomb heuristic is optimal~\cite{Zemel:84}, yielding 
\begin{equation}
\label{eq:dvi} \lim_{m \to \infty} d_{\mathcal{V}_i}m^{1/2} = \sqrt{\frac{2\,\mathcal{A}}{3\sqrt{3}}} \approx 0.62 \sqrt{\mathcal{A}}, \quad \forall i \in \{1, \ldots, m\}.\end{equation}
Clearly, in these conditions the first term in \eqref{eq:bound_acc} is 
the dominant one. 

Another consequence of the optimality of the honeycomb heuristic is that 
\begin{equation}
\label{eq:hstarhex}
\lim_{m \to \infty} \mathcal{H}^*_m(\mathcal{Q}) m^{1/2}= c_\mathrm{hex} \sqrt{\mathcal{A}}.
\end{equation}
Integrating~\eqref{eq:bound_acc} over $\mathcal{Q}$, multiplying both sides by $m^{1/4}$, and taking the limit as $m \to \infty$, 
we get 
$$  \liminf_{m \to \infty} \TMSDI m^{1/4}  \ge 
\liminf_{m \to \infty} \frac{\mathcal{H}^*_m(\mathcal{Q}) m^{1/2}}{\sqrt{2 \umax  d_{\mathcal{V}_i}m^{1/2}}}. 
$$
Finally, using \eqref{eq:dvi} and \eqref{eq:hstarhex}, we get the desired result.

We will only sketch the proof for the DD case. The minimum travel time for a DD robot can be decomposed into the sum of the cost of turning towards the target point, plus the Euclidean distance between the robot and the target point.  The Euclidean term  vanishes as $m$ increases. The turning cost on the other hand remains bounded away from zero. Since the robot's initial heading is chosen randomly, the expected turning angle is $\pi/4$, which combined with the maximum turning rate $\wmax/\rho$ yields the stated result.
\end{proof}

In other words, for DI robots, the MS algorithm requires them to stay stationary in absence of any outstanding service requests. Once a service request is assigned to a robot, the amount of time spend in attaining the maximum speed $\vmax$ becomes significant as the location of assigned service requests start getting closer. Similar arguments hold for DD robots. 

An alternate approach, as proposed in the next algorithm, is to keep the robots moving rather than waiting in absence of outstanding service requests. The algorithm assigns dynamic regions of responsibility to the robots.  

\subsection*{The Strip Loitering (SL) Algorithm}
This algorithm is an adaptation of a similar algorithm proposed in \cite{Enright.Savla.ea:CDC08} for Dubins vehicle, i.e., vehicles constrained to move forward with constant speeds along paths of bounded curvature. 

\begin{figure}[htb]
  \centerline{\includegraphics[angle=90,width=0.9\linewidth]{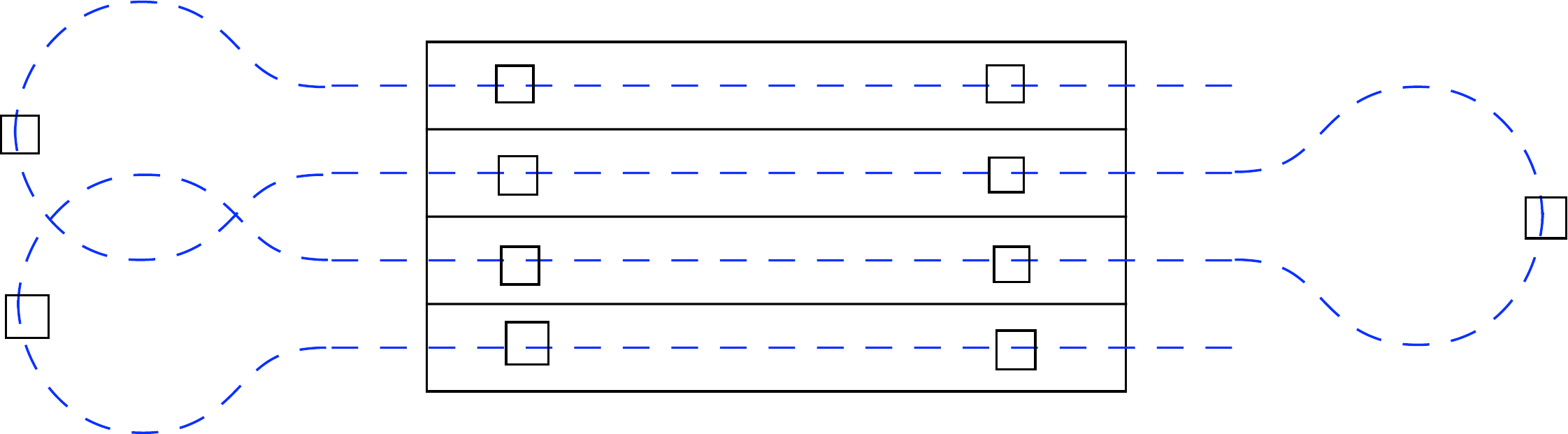}}
\caption{Depiction of the loitering path for the double integrator robots. The segment providing closure of the loitering path (returning the robots from the end of the last strip to the beginning of the first strip) is not shown here for clarity of the drawing.}
\label{fig:SL}
\end{figure}

\begin{figure}[htb]
  \centerline{\includegraphics[width=0.5\linewidth]{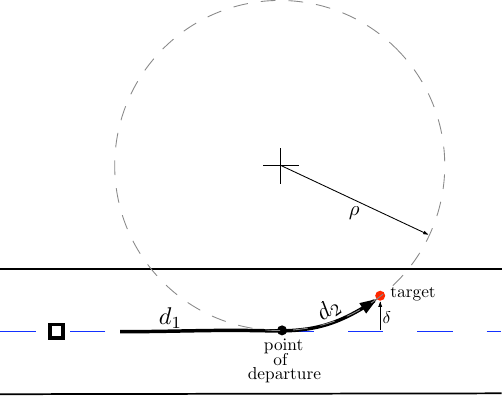}}
\caption{Close-up of the loitering path with construction of the point of departure and the distances $\delta$, $d_1$, and $d_2$ for a given target, at the instant of appearance.}
\label{fig:SLclose}
\end{figure}

Let the robots move with constant speed $\vstar=\min\{\vmax,\frac{\sqrt{\sqrt{\area}\umax}}{3.22}\}$ and follow a loitering path which is constructed as follows. Divide $\domain$ into strips of width $w$ where $w=\min\Big\{\Big(\frac{4}{3
    \sqrt{\rhostar}}\frac{\area+10.38 \rhostar \sqrt{\area}}{m}\Big)^{2/3},2
  \rhostar \Big\}$, where $\rhostar:=\frac{{\vstar}^2}{\umax}$. Orient the strips along a side of $\domain$.  Construct a closed path which can be traversed by a double integrator robot while always moving with constant speed $\vstar$. This closed path runs along the longitudinal bisector of
each strip, visiting all strips from top-to-bottom, making U-turns
between strips at the edges of $\domain$, and finally returning to the
initial configuration. The $m$ robots loiter on this path, equally
spaced, in terms of path length. A depiction of the Strip Loitering algorithm can be viewed in
Figure~\ref{fig:SL}.  Moreover, in Figure~\ref{fig:SLclose} we define
two distances that are important in the analysis of this algorithm.
Variable $d_2$ is the length of the shortest path departing from the
loitering path and ending at the target (a circular arc of radius
$\rhostar$).  The robot responsible for	
visiting the target is the one closest in terms of loitering path
length (variable $d_1$) to the point of departure, at the time of target-arrival.  Note that there may be robots closer to the target in terms of the actual distance.  However, we find that the
assignment strategy described above lends itself to tractable
analysis.

After a robot has serviced a target, it must return to its place in the
loitering path.  We now describe a method to accomplish this task
through the example shown in Fig. \ref{fig:SLclose}.  After making a
left turn of length $d_2$ to service the target, the robot makes a right
turn of length $2d_2$ followed by another left turn of length $d_2$,
returning it to the loitering path.  However, the robot has fallen
behind in the loitering pattern.  To rectify this, as it nears the end
of the current strip, it takes its U-turn early.

Let $\TSLDI$ be the coverage cost as given by the above algorithm for DI robots. We now state an upper bound on $\TSLDI$.

\begin{theorem}[Analysis of the SL algorithm]
\label{thm:SL-bound}
The coverage cost for a team of DI robots implementing the SL algorithm satisfies the following asymptotic upper bound.
\begin{equation*}
\limsup_{m \to +\infty} \TSLDI  m^{1/3} \leq \frac{1.238}{\vstar} \big(\rhostar \area+10.38 {\rhostar}^2 \sqrt{\area} \big)^{1/3}.
\end{equation*} 
\end{theorem} 
\begin{proof}
Since a similar algorithm for Dubins vehicle was analyzed in \cite{Enright.Savla.ea:CDC08}, we only outline the proof here and refer to \cite{Enright.Savla.ea:CDC08} for more details.
Denote the length of the closed path as $L_1$. Due to equal
spacing of the robots along the loitering path,
\begin{equation} 
  \label{eq:d_1}
  \mathrm{E}[d_1] = \frac{L_1}{2m}.
\end{equation}
We now calculate an upper bound on $L_1$. To that effect, let
$N_{\mathrm{strips}}$ be the number of strips, $L_{\mathrm{strip}}$ be
the length travelled along a single strip, $L_{\mathrm{u-turn}}$ be
the length of a u-turn and $L_{\mathrm{closure}}$ be the length of the
closure path. With these notations, $L_1$ can be bounded as
\begin{equation}
\label{eq:L_1}
L_1 \leq  N_{\mathrm{strips}} L_{\mathrm{strip}} + (N_{\mathrm{strips}}-1) L_{\mathrm{u-turn}} + L_{\mathrm{closure}}.
\end{equation}
One can compute bounds for various terms on the right side of Equation~\eqref{eq:L_1}.
 Substituting these bounds into
Equation~\eqref{eq:L_1} and taking into account
Equation~\eqref{eq:d_1} we get that
\begin{equation}
  \label{eq:d_1-final}
  \mathrm{E}[d_1] \leq \frac{\area+10.38 \rhostar \sqrt{\area}}{2mw}+ \frac{2 \sqrt{\area} + 6.19 \rhostar }{m}.
\end{equation}
To calculate $\mathrm{E}[d_2]$ we define
$\delta$ as the smallest distance from the target to any point on the
loitering path (see Fig. \ref{fig:SLclose}). Since $d_2(s)=2 \rhostar
\sin^{-1}(\sqrt{\frac{s}{2 \rhostar}})$ for $s \le \rhostar$ and $\delta$ is
uniformly distributed between $0$ and $w/2$,
\begin{equation}
\label{eq:d_2}
\mathrm{E}[d_2]=\frac{4 \rhostar}{w} \int_{0}^{w/2} \sin^{-1}\left(\sqrt{\frac{s}{2 \rhostar}}\right) ds \leq \frac{3}{4} \sqrt{\rhostar w}.
\end{equation}
 
 Therefore, the coverage cost is given by
\begin{equation}
\label{eq:d1d2}
\overline T_\mathrm{SL}  \le \frac{\mathrm{E}[d_1]+ \mathrm{E}[d_2]}{\vstar}.
\end{equation}
Therefore, from Equations~\eqref{eq:d1d2}, \eqref{eq:d_1-final} and
\eqref{eq:d_2} we get that for $w \leq 2 \rho$,
\begin{equation}
  \label{eq:d1d2-final}
  \TSLDI  \le \frac{\area+10.38 \rhostar \sqrt{\area}}{2mw\vstar}+ \frac{2 \sqrt{\area} + 6.19 \rhostar }{m\vstar} + \frac{3}{4\vstar} \sqrt{\rhostar w}.
\end{equation}
In order to get the least upper bound, we minimize the right hand side
of Equation~\eqref{eq:d1d2-final} with respect to $w$ subject to the
constraint that $w \leq 2 \rhostar$. 
 
\end{proof}

\begin{remark}
Theorem~\ref{thm:SL-bound} and Theorem~\ref{thm:asymp-lower-bound} imply that $\ToptDI$ belongs to $\Theta(1/m^{1/3})$. Moreover, Theorem~\ref{thm:SL-bound} and Theorem~\ref{thm:tighter_lb} together with Equation~\ref{eq:weber-bounds} imply that $\TSLDI/\TMSDI \to 0^+$ as $m \to +\infty$. Hence, asymptotically, the SL algorithm outperforms the MS algorithm and is within a constant factor of the optimal.
\end{remark}

We now present the second algorithm for DD robots.

\subsection*{The Median Clustering (MC) Algorithm}
\begin{figure}[htb]
  \centerline{\includegraphics[width=0.4\linewidth]{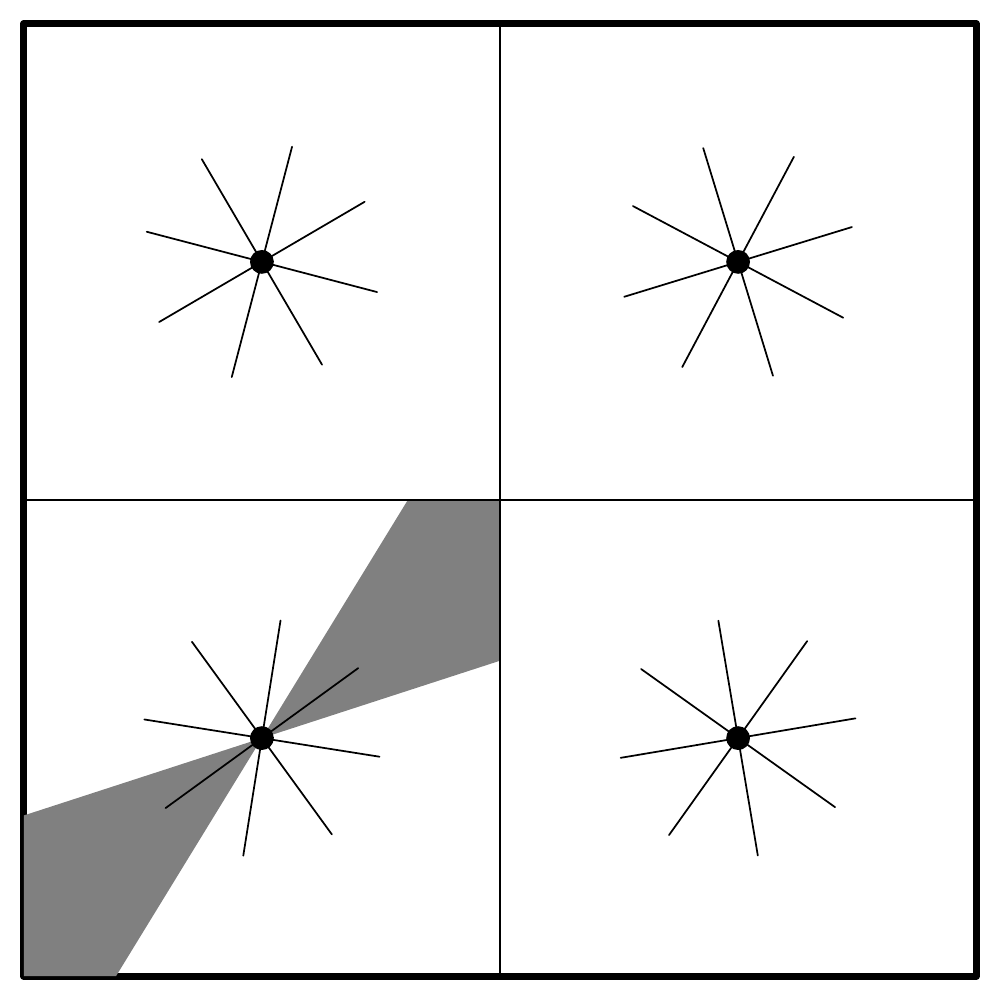}}
\caption{Depiction of the Median Clustering Algorithm with teams of 4 robots each. The shaded region represents a typical \emph{region of responsibility} for a robot.}
\label{fig:mediancluster}
\end{figure}

Form as many teams of robots with $k:=\left\lceil 4.09 \Big(\frac{\rho}{\sqrt{A}} \Big)^{2/3} m^{1/3} \right\rceil$ robots in each team. If there are additional robots, group them into one of these teams. Let $n:=\left\lfloor \frac{m}{k} \right\rfloor$ denote the total number of teams formed. Position these $n$ teams at the $n$-median locations of $\domain$, i.e., all the robots in a team are co-located at the median location of its team. Within each team $j$, $j \in \until{n}$,  the headings of the robots belonging to that team are selected as follows. Let $\ell_j \ge k$ be the number of robots in team $j$. Pick a direction randomly. The heading of 
one robot is aligned with this direction. The heading of the second robot is selected to be along a line making an angle $\frac{\pi}{\ell}$, in the counter-clockwise direction, with the first robot. The headings of the remaining robots are selected similarly along directions making $\frac{\pi}{\ell}$-angle with the previous one (see Figure~\ref{fig:mediancluster}). These headings will be called the \emph{median headings} of the robots. Each robot in a team is assigned a \emph{dominance region} which is the region formed by the intersection of double cone making half angle of $\frac{\pi}{2\ell}$ with its median heading and the Voronoi cell belonging to the team (see Figure~\ref{fig:mediancluster}). When a service request appears, it is assigned to the robot whose dominance region contains its location. The assigned robot travels to the service location in the fastest possible way and, upon the completion of the service, returns to the median location of its team and aligns itself with its original median heading.

Let $\TMCDD$ be the coverage cost as given by the above policy. We now state an upper bound on $\TMCDD$ in the following theorem.

\begin{theorem}[Analysis of the MC algorithm]
\label{thm:MC-bound}
The coverage cost for a team of DD robots while implementing the MC algorithm satisfies the following asymptotic upper bound.
\begin{equation*}
\limsup_{m \to +\infty} \TMCDD m^{1/3} \leq \frac{1.15}{\wmax} (\rho \area)^{1/3}.
\end{equation*} 
\end{theorem}
\begin{proof}
The travel time for any robot from its median location $p$ to the location $q$ of a service request is upper bounded by $\frac{\|p-q\|}{\wmax}+\frac{\pi \rho}{2 \wmax k}$. Taking the expected value of this quantity while taking into consideration the assignment policy of the service requests gives us that
\begin{equation}
\label{eq:TMCDD-step1}
\TMCDD \leq \frac{\weberfunc{n}{\domain}}{\wmax}+\frac{\pi \rho}{2 \wmax k}.
\end{equation}
From Equation~\eqref{eq:weber-bounds}, $\weberfunc{n}{\domain} \leq 0.38 \sqrt{\frac{\area}{n}}$. Moreover, for large $m$, $n \approx m/k$. This combined with Equation~\eqref{eq:TMCDD-step1}, one can write that, for large $m$,
\begin{equation}
\label{eq:TMCDD-step2}
\TMCDD \leq \frac{0.38}{\wmax}\sqrt{\frac{\area}{m/k}}+\frac{\pi \rho}{2 \wmax k}.
\end{equation}
The right hand side of Equation~\eqref{eq:TMCDD-step2} is minimized when $k=4.09 \Big(\frac{\rho}{\sqrt{A}} \Big)^{2/3} m^{1/3}$. Substituting this into Equation~\eqref{eq:TMCDD-step2}, one arrives at the result.
\end{proof}

\begin{remark}
Theorem~\ref{thm:MC-bound} and Theorem~\ref{thm:asymp-lower-bound} imply that $\ToptDD$ belongs to $\Theta(1/m^{1/3})$. Moreover, Theorem~\ref{thm:MC-bound} and Theorem~\ref{thm:tighter_lb} together with Equation~\ref{eq:weber-bounds} imply that $\TMCDD/\TMSDD \to 0^+$ as $m \to +\infty$. Hence, asymptotically, the MC algorithm outperforms the MS algorithm and is within a constant factor of the optimal.
\end{remark}

\section{Conclusion}
In this paper, we considered a coverage problem for a mobile robotic network modeled as double integrators and differential drives. We observe that the optimal algorithm for omni-directional robots is a reasonable solution for sparse networks of double integrator or differential drive robots. However, these algorithms do not perform well for large networks because they don't take into consideration the effect of motion constraints. We propose novel algorithms that are within a constant factor of the optimal for the DI as well as DD robots and prove that the coverage cost for both of these robots is of the order $1/m^{1/3}$.

In future, we would like to obtain sharper bounds on the coverage cost so that we can make meaningful predictions of the onset of reconfiguration in terms of system parameters. It would be interesting to study the problem for non-uniform distribution of targets and for higher intensity of arrival. Also, this research opens up possibilities of reconfiguration due to other constraints, like sensors (isotropic versus anisotropic), type of service requests (distributable versus in-distributable), etc. Lastly, we plan to apply this research in understanding phase transition in naturally occurring systems, e.g., desert locusts~\cite{Buhl.Sumpter.ea:06}. 



\bibliographystyle{abbrv} 
\bibliography{frazzoli,efmain,alias}

%


\end{document}